\documentclass{article}

\usepackage{arxiv}

\usepackage[utf8]{inputenc} 
\usepackage[T1]{fontenc}    
\usepackage{amsmath,amssymb,amsfonts}%
\usepackage{amsthm}%
\usepackage{algorithm}%
\usepackage{algorithmicx}%
\usepackage{algpseudocode}%
\usepackage{multirow}
\usepackage{hyperref}       
\usepackage{cleveref}
\usepackage{url}            
\usepackage{booktabs}       
\usepackage{amsfonts}       
\usepackage{nicefrac}       
\usepackage{microtype}      
\usepackage{lipsum}		
\usepackage{graphicx}
\usepackage{xcolor}
\usepackage[numbers]{natbib}
\usepackage{doi}
\usepackage[normalem]{ulem}

\title{FC-KAN: Function Combinations in Kolmogorov-Arnold Networks}

\author{ \href{https://orcid.org/0000-0003-0321-5106}{\includegraphics[scale=0.06]{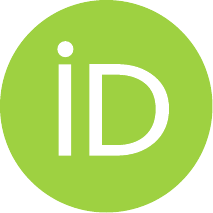}\hspace{1mm}Hoang-Thang Ta}
	\\ Department of Information Technology, Dalat University\\
	Lam Dong, Vietnam \\
	\texttt{thangth@dlu.edu.vn} \\
	   \AND
	\href{https://orcid.org/0000-0002-3991-2037}{\includegraphics[scale=0.06]{orcid.pdf}\hspace{1mm}Duy-Quy Thai} \\
	  Department of Information Technology, Dalat University \\
	Lam Dong, Vietnam \\
	\texttt{quytd@dlu.edu.vn} \\
     \AND
	\href{https://orcid.org/0000-0002-8581-0891}{\includegraphics[scale=0.06]{orcid.pdf}\hspace{1mm}Abu Bakar Siddiqur Rahman} \\
	  College of Information Science and Technology, University of Nebraska Omaha \\
	Nebraska, USA \\
	  \texttt{abubakarsiddiqurra@unomaha.edu} \\
     \AND
	\href{https://orcid.org/0000-0003-3901-3522}{\includegraphics[scale=0.06]{orcid.pdf}\hspace{1mm}Grigori Sidorov} \\
	  Centro de Investigación en Computación (CIC), Instituto Politécnico Nacional (IPN) \\
	CDMX, Mexico \\
	  \texttt{grigori@cic.ipn.mx} \\
	 \AND
	\href{https://orcid.org/0000-0001-7845-9039}{\includegraphics[scale=0.06]{orcid.pdf}\hspace{1mm}Alexander Gelbukh} \\
	Centro de Investigación en Computación (CIC), Instituto Politécnico Nacional (IPN) \\
	 CDMX, Mexico \\
	 \texttt{gelbukh@cic.ipn.mx} \\
}

\renewcommand{\shorttitle}{\textit{arXiv} Template}

\hypersetup{
pdftitle={A template for the arxiv style},
pdfsubject={q-bio.NC, q-bio.QM},
pdfauthor={David S.~Hippocampus, Elias D.~Striatum},
pdfkeywords={First keyword, Second keyword, More},
}

\begin{document}
\maketitle

\begin{abstract}
In this paper, we introduce FC-KAN, a Kolmogorov-Arnold Network (KAN) that leverages combinations of popular mathematical functions such as B-splines, wavelets, and radial basis functions on low-dimensional data through element-wise operations. We explore several methods for combining the outputs of these functions, including sum, element-wise product, the addition of sum and element-wise product, representations of quadratic and cubic functions, concatenation, linear transformation of the concatenated output, and others. In our experiments, we compare FC-KAN with a multi-layer perceptron network (MLP) and other existing KANs, such as BSRBF-KAN, EfficientKAN, FastKAN, and FasterKAN, on the MNIST and Fashion-MNIST datasets. Two variants of FC-KAN, which use a combination of outputs from B-splines and \textcolor{blue}{Derivative of Gaussians} (DoG) and from B-splines and linear transformations in the form of a quadratic function, outperformed overall other models on the average of 5 independent training runs. We expect that FC-KAN can leverage function combinations to design future KANs. Our repository is publicly available at: \url{https://github.com/hoangthangta/FC_KAN}.

\textcolor{blue}{Please refer to the journal version at: \url{https://www.sciencedirect.com/science/article/abs/pii/S0020025526000344}}

*Note: The function used in this work is \textcolor{blue}{Derivative of Gaussian} (DoG), not \textcolor{red}{\sout{Difference of Gaussians}}. Unfortunately, we cannot correct this in the journal version.
\end{abstract}


\section{Introduction}

Recently, the works by \citet{liu2024kan,liu2024kan2.0} have gained significant attention from the research community for applying the Kolmogorov-Arnold representation theorem (KART) in neural networks through the introduction of KANs. They highlighted the use of learnable activation functions as "edges" to fit training data better, in contrast to the traditional use of fixed activation functions as "nodes" in multi-layer perceptrons (MLPs). The foundation of KANs is based on KART, which was developed to address Hilbert's 13th problem~\cite{sternfeld2006hilbert}, specifically, the statement, ``\textit{Prove that the equation of seventh degree $x^7 + ax^3 + bx^2 + cx + 1 = 0$ is not solvable by means of any continuous functions of only two variables}''. KART asserts that any continuous function of multiple variables can be represented as a sum of continuous functions of a single variable~\cite{kolmogorov1957representation}.  


Inspired by KANs~\cite{liu2024kan,liu2024kan2.0}, numerous researchers have explored developing novel neural networks using widely known polynomials and basis functions. While most works use a single function in constructing KANs~\cite{li2024kolmogorov,athanasios2024,bozorgasl2024wav,ss2024chebyshev}, a few have investigated function combinations~\cite{ta2024bsrbf,yang2024activation,altarabichi2024rethinking,liu2024kan2.0}. In several works on MLP-based neural networks, function combinations typically appear in activation functions to enhance model performance and stability~\cite{jie2021regularized,xu2020comparison,zhang2015genetic}. In KANs, the combinations occur directly on the basis functions used to fit input data, rather than through activation functions. For instance, \citet{ta2024bsrbf} designed BSRBF-KAN, which combines B-splines and Gaussian Radial Basis Functions (GRBFs) in all network layers. The authors only applied addition operations to data tensors without incorporating element-wise multiplications, which we believe does not effectively capture data features. Although \citet{liu2024kan2.0} used element-wise multiplications alongside additions in MultKAN to enhance data capture, their focus was limited to small-scale examples. Another study developed ReLU-KAN by replacing B-splines with ReLU activations, along with addition and multiplication operations, to improve GPU parallelization and data fitting~\cite{qiu2024relu}, but it focused solely on ReLU without exploring combinations with other functions.

With the purpose of exploiting data features efficiently, we propose a novel KAN, FC-KAN (\textbf{F}unction \textbf{C}ombinations in \textbf{K}olmogorov-\textbf{A}rnold \textbf{N}etworks), which leverages various functions to capture input data throughout the network layers and combines them in low-dimensional spaces, such as the output layer, using various methods mainly based on element-wise operations, including sum, product, the combination of sum and product, representations of quadratic and cubic functions, and concatenation. We avoid using a higher-degree function due to their increased computational demands on data tensors, which can lead to memory errors. As a result, FC-KAN is able to capture more data features, leading to improved performance on the MNIST and Fashion-MNIST datasets compared to other KAN networks. Moreover, employing n-degree functions aligns with the core concept of KAN, where they are used both to capture input data and to represent data features in the output. We expect this exciting development to drive the proliferation of function combinations in neural networks, particularly in KANs. In summary, our main contributions are:

\begin{itemize}
    \item Introduce FC-KAN, a novel KAN that explores function combinations through various methods applied to low-dimensional data.
    \item Evaluate the performance of different combination methods in FC-KAN on two image classification datasets: MNIST and Fashion-MNIST.
    \item Investigate whether model performance in full-data training can be inferred from models trained with limited data.
\end{itemize}

Aside from this section, the paper is organized as follows: Section 2 discusses related work on KART and KANs. Section 3 details our methodology, covering KART, the design of the KAN architecture, several existing KANs, and FC-KAN. Section 4 presents our experiments, comparing FC-KAN variants with MLP and other existing KANs using data from the MNIST and Fashion-MNIST datasets. Besides, this section includes a comparison of combination methods within FC-KAN, a misclassification analysis, and an analysis of model performance with limited data. We mention some limitations of our study in Section 5. Lastly, Section 6 offers our conclusions and potential directions for future research.

\section{Related Works}\label{sec_related_works}

\subsection{KART and KAN}
In 1957, Kolmogorov provided a proof to Hilbert’s 13th problem by showing that any multivariate continuous function can be expressed as a combination of single-variable functions and additions, a concept known as KART~\cite{kolmogorov1957representation,braun2009constructive}. This theorem has been utilized in many studies to develop neural networks~\cite{zhou2022treedrnet,sprecher2002space,koppen2002training,lin1993realization,leni2013kolmogorov,lai2021kolmogorov}. However, there is an ongoing debate about the applicability of KART in neural network design. \citet{girosi1989representation} argued that KART's relevance to neural networks is questionable because the inner function $\phi_{q,p}$ in \Cref{eq:kart} may be highly non-smooth~\cite{vitushkin1954hilbert}, which could hinder $f$ from being smooth—a key attribute for generalization and noise resistance in neural networks. Conversely, \citet{kuurkova1991kolmogorov} contended that KART is applicable to neural networks, showing that linear combinations of affine functions can effectively approximate all single-variable functions using certain sigmoidal functions.

Despite the long history of KART's application in neural networks, it had not garnered significant attention in the research community until the recent study by \citet{liu2024kan}. They suggested moving away from strict adherence to KART and generalizing it to develop KANs with additional neurons and layers. Our intuition aligns with this perspective as it helps to mitigate the issue of non-smooth functions when applying KART to neural networks. Consequently, KANs have the potential to outperform MLPs in both accuracy and interpretability for small-scale AI + Science tasks. However, KANs have also faced criticism from \citet{dhiman2024kan}, who argue that they are essentially MLPs with spline-based activation functions, in contrast to traditional MLPs with fixed activation functions. KANs also face the problem of using too many parameters compared to MLPs. \citet{yu2024kan} indicated that KANs are not better than MLPs when using the same number of parameters and FLOPs, except for symbolic formula representation tasks.

By introducing a new perspective to the scientific community in the neural network designs, KANs inspired many works to prove their effectiveness by topics, including expensive problems~\cite{hao2024first}, keyword spotting~\cite{xu2024effective}, mechanics problems~\cite{abueidda2024deepokan}, quantum computing~\cite{kundu2024kanqas,wakaura2024variational,troy2024sparks}, survival analysis~\cite{knottenbelt2024coxkan}, time series forecasting~\cite{genet2024tkan,xu2024kolmogorov,vaca2024kolmogorov,genet2024temporal,han2024kan4tsf}, and vision tasks~\cite{li2024u,cheon2024demonstrating,ge2024tc}. Also, many novel KANs utilize well-known mathematical functions, particularly those capable of handling curves, such as B-splines~\cite{de1972calculating} (Original KAN~\cite{liu2024kan}, EfficientKAN\footnote{https://github.com/Blealtan/efficient-kan}, BSRBF-KAN~\cite{ta2024bsrbf}), Gaussian Radial Basis Functions (GRBFs) (FastKAN~\cite{li2024kolmogorov}, DeepOKAN~\cite{abueidda2024deepokan}, and BSRBF-KAN~\cite{ta2024bsrbf}),  Reflection SWitch Activation Function (RSWAF) in FasterKAN~\cite{athanasios2024}, Chebyshev polynomials (TorchKAN~\cite{torchkan}, Chebyshev KAN~\cite{ss2024chebyshev}), Legendre polynomials (TorchKAN~\cite{torchkan}), Fourier transform (FourierKAN\footnote{https://github.com/GistNoesis/FourierKAN/}, FourierKAN-GCF~\cite{xu2024fourierkan}), wavelets~\cite{bozorgasl2024wav,seydi2024unveiling}, and other polynomial functions~\cite{teymoor2024exploring}. 

\subsection{Function Combinations in KANs and Other Neural networks}

Several works utilize the function combinations to design novel KANs. \citet{ta2024bsrbf} mentioned the combination of functions -- B-splines and radial basis functions -- in designing KANs. Their BSRBF-KAN showed better convergence on the training data for MNIST and Fashion-MNIST. \citet{liu2024kan2.0} introduced MultKAN, which consists of multiplication operations to capture multiplicative structures in data better. Unlike KAN~\cite{liu2024kan}, which directly copies nodes, MultKAN uses both addition nodes (copied from subnodes) and multiplication nodes (multiplying multiple subnodes). However, their work focused on small-scale examples only. \citet{yang2024activation} utilized function combinations to create optimal activation functions at each node using an adaptive strategy, addressing the drawbacks of single activation functions in their S-KAN model. They also extended S-KAN to S-ConvKAN, which showed superior performance in image classification tasks, outperforming CNNs and KANs with comparable structures. In another work, ~\citet{altarabichi2024rethinking} proposed the replacement of the sum with the average function in KAN neurons that can improve the model performance and keep the training stability in their DropKAN~\cite{altarabichi2024dropkan}. Unlike other studies, \citet{qiu2024relu} developed ReLU-KAN, replacing B-splines with a novel basis function that leverages matrix operations (addition and multiplication) and ReLU activations to enhance GPU parallelization and fitting performance.

Before the appearance of KANs, several studies focused on constructing combinations of activation functions or utilizing a set of flexible activations in neural networks to enhance model performance and stability. \citet{jie2021regularized} introduced a new family of flexible activation functions for LSTM cells, along with another family developed by combining ReLUs and ELUs. Their findings demonstrate that LSTM models using the P-Sig-Ramp flexible activations significantly improve time series forecasting. Additionally, the P-E2-ReLU activation exhibits enhanced stability and performance in lossy image compression tasks with convolutional autoencoders. \citet{xu2020comparison} investigated a selection of widely used activation functions across various datasets in classification and regression tasks. They then created combinations of the best-performing activation functions within a convex restriction, showing improved performance compared to the corresponding base activation functions in a standard broad learning system. In another study, \citet{zhang2015genetic} developed a novel deep neural network (DNN) that optimizes activation function combinations for different neurons based on extensive simulations. The experiments showed that their DNNs outperformed those that rely on a single activation function. Instead of using the inner product between data tensors and weights, ~\citet{fan2020universal} replaced this operation with a quadratic function in neurons within deep quadratic neural networks. Quadratic neurons provide enhanced expressive capability compared to conventional neurons, highlighting the advantages of quadratic networks in terms of expressive efficiency, unique representation, compact architecture, and computational capacity.








Some other works focus on combining function outputs using tensor operations, exploring various methods to aggregate outputs efficiently. When the data dimensions are the same, element-wise operations such as summation or product can be applied to combine these outputs~\cite{ji2019survey}, which appear in many multi-modal tasks~\cite{fukui2016multimodal,yu2017multi,wu2018multi}. However, the computational challenges posed by high-dimensional data, such as inefficiencies in tensor-product operations, have led to a growing body of work aimed at optimizing and accelerating these processes~\cite{cui2024acceleration,zachariadis2020accelerating}. Furthermore, other studies have focused on tensor fusion and tensor decomposition techniques to extract meaningful features from data tensors and to simplify data combinations~\cite{zadeh2017tensor,oseledets2011tensor,liu2018efficient,jin2020dual}.

\section{Methodology}
\label{sec:methodology}

\subsection{Kolmogorov-Arnold Representation Theorem}

A KAN is based on KART, which states that any continuous multivariate function $f$ defined on a bounded domain can be represented as a finite combination of continuous single-variable functions and their additions~\cite{chernov2020gaussian,schmidt2021kolmogorov}. \textcolor{blue}{For a set of variables $\mathbf{x} = (x_1, x_2, \ldots, x_n) \in [0,1]^n$, 
let $f : [0,1]^n \to \mathbb{R}$ be a continuous multivariate function. Then $f$ can be expressed as:}~\cite{ismailov2025addressing}

\begin{equation}
\begin{aligned}
f(\mathbf{x}) = f(x_1, \ldots, x_n) = \sum_{q=1}^{2n+1} \Phi_q \left( \sum_{p=1}^{n} \phi_{q,p}(x_p) \right) 
\end{aligned}
\label{eq:kart}
\end{equation}
which has two types of summations: the outer sum and the inner sum. The outer sum, $\sum_{q=1}^{2n+1}$, aggregates $2n+1$ terms of $\Phi_q$ ($\mathbb{R} \to \mathbb{R}$). The inner sum, on the other hand, aggregates $n$ terms for each $q$, where each term $\phi_{q,p}$ ($\phi_{q,p} \colon [0,1] \to \mathbb{R}$) denotes a continuous function of a single variable $x_p$.

\subsection{Design of KANs}
\label{KAN_design}

Remind an MLP that consists of affine transformations and non-linear functions. Starting with an input $\mathbf{x}$, the network processes it through a series of weight matrices across layers (from layer $0$ to layer $L-1$) and applies the non-linear activation function $\sigma$ to produce the final output.

\begin{equation}
\begin{aligned}
\text{MLP}(\mathbf{x}) &= (W_{L-1} \circ \sigma \circ W_{L-2} \circ \sigma \circ \cdots \circ W_1 \circ \sigma \circ W_0) \mathbf{x} \\
&= \sigma \left( W_{L-1} \sigma \left( W_{L-2} \sigma \left( \cdots \sigma \left( W_1 \sigma \left( W_0 \mathbf{x} \right) \right) \right) \right) \right)
\end{aligned}
\label{eq:mlp}
\end{equation}

Inspired by KART, \citet{liu2024kan} developed KANs but recommended extending the approach to incorporate greater network widths and depths. To address this, appropriate functions $\Phi_q$ and $\phi_{q,p}$ in \Cref{eq:kart} need to be identified. A typical KAN network with $L$ layers processes the input $\mathbf{x}$ to produce the output as follows:

\begin{equation}
\begin{aligned}
\text{KAN}(\mathbf{x}) = (\Phi_{L-1} \circ \Phi_{L-2} \circ \cdots \circ \Phi_1 \circ \Phi_0)\mathbf{x}
\end{aligned}
\label{eq:kan}
\end{equation}
which $\Phi_{l}$ is the function matrix of the $l^{th}$ KAN layer or a set of pre-activations. Let denote the neuron $i^{th}$ of the layer $l^{th}$ and the neuron $j^{th}$ of the layer $l+1^{th}$. The activation function $\phi_{l,i,j}$ connects $(l, i)$ to $(l + 1, j)$:

\begin{equation}
\begin{aligned}
\phi_{l,j,i}, \quad l = 0, \cdots, L - 1, \quad i = 1, \cdots, n_l, \quad j = 1, \cdots, n_{l+1}
\end{aligned}
\label{eq:acti_funct}
\end{equation}
with $n_l$ is the number of nodes of the layer $l^{th}$. So now, the function matrix $\Phi_{l}$ can be represented as a matrix $n_{l+1} \times n_{l}$ of activations as:

\begin{equation}
\begin{aligned}
\mathbf{x}_{l+1} = 
\underbrace{\left(
\begin{array}{cccc}
\phi_{l,1,1}(\cdot) & \phi_{l,1,2}(\cdot) & \cdots & \phi_{l,1,n_l}(\cdot) \\
\phi_{l,2,1}(\cdot) & \phi_{l,2,2}(\cdot) & \cdots & \phi_{l,2,n_l}(\cdot) \\
\vdots & \vdots & \ddots & \vdots \\
\phi_{l,n_{l+1},1}(\cdot) & \phi_{l,n_{l+1},2}(\cdot) & \cdots & \phi_{l,n_{l+1},n_l}(\cdot)
\end{array}\right)}_{\Phi_{l}} \mathbf{x}_l
\label{eq:function_matrix}
\end{aligned}
\end{equation}

\subsection{Implementation of the Current KANs}


\begin{figure*}[htbp]
  \centering
\includegraphics[scale=0.95]{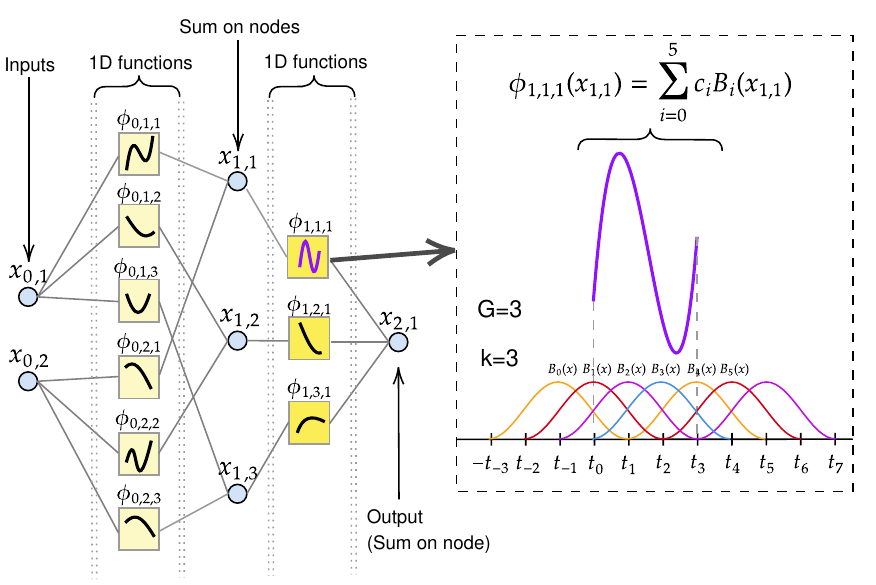}
  \centering
  \caption{Left: The structure of KAN(2,3,1). Right: The simulation of how to calculate $\phi_{1,1,1}$ by control points and B-splines. \(G\) and \(k\) is the grid size and the spline order, the number of B-splines equals \(G + k = 3 + 3 = 6\).}
\label{fig:kan_diagram}
\end{figure*}

\begin{figure*}[htbp]
  \centering
\includegraphics[scale=0.6]{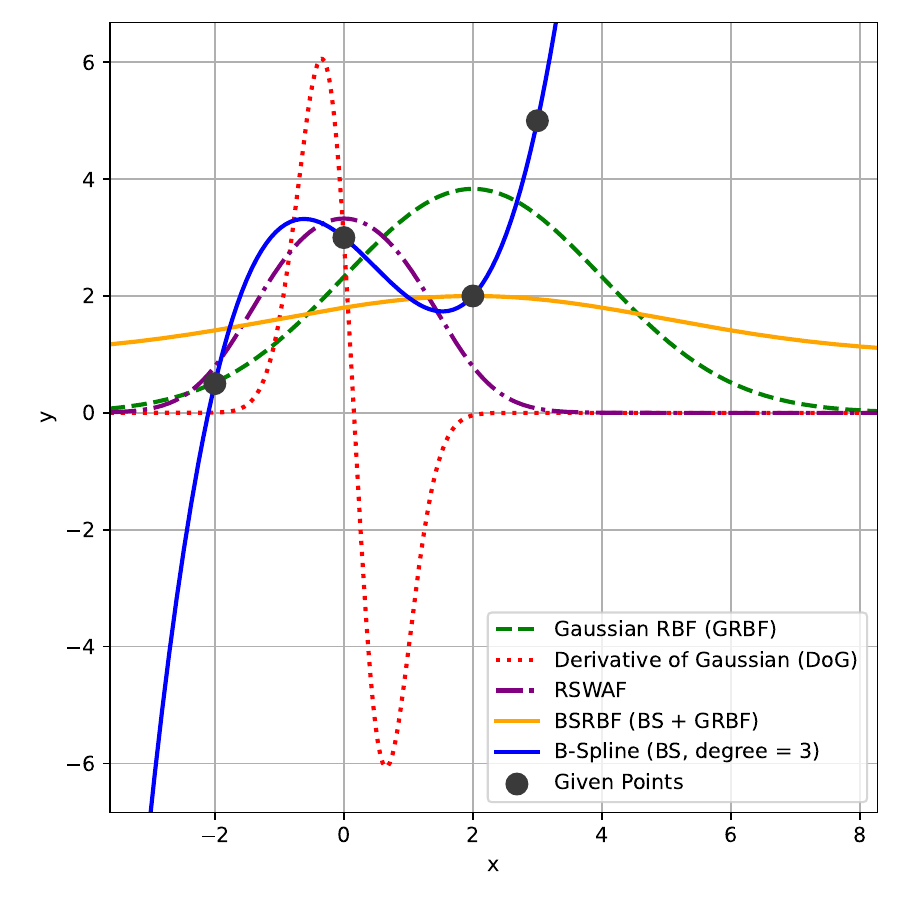}
  \centering
  \caption{The plots of the functions when fitted to pass through the 4 chosen points.}
\label{fig:func_plot}
\end{figure*}

\textbf{LiuKAN}\footnote{We refer to the original KAN as LiuKAN, following the first author's last name~\cite{liu2024kan}, while another work~\cite{bozorgasl2024wav} refers to it as Spl-KAN.} was created by \citet{liu2024kan} by using the residual activation function $\phi(x)$ as the sum of the base function and the spline function with their corresponding weight matrices $w_b$ and $w_s$:

\begin{equation}
\begin{aligned}
\phi(x) = w_b b(x) + w_s spline(x)
\end{aligned}
\label{eq:acti_funct_imp}
\end{equation}

\begin{equation}
\begin{aligned}
b(x) = silu(x) = \frac{x}{1 + e^{-x}}
\end{aligned}
\label{eq:b_function}
\end{equation}

\begin{equation}
\begin{aligned}
spline(x) = \sum_{i}c_iB_i(x)
\end{aligned}
\label{eq:spline_function}
\end{equation}
where \(b(x)\) equals to $silu(x)$ as in \Cref{eq:b_function} and \(spline(x)\) is expressed as a linear combination of B-splines $B_i$s and control points (coefficients) $c_i$s as in \Cref{eq:spline_function}. Each activation function is activated with \(w_s = 1\) and \(spline(x) \approx 0\), while \(w_b\) is initialized by using Xavier initialization. 

\Cref{fig:kan_diagram} shows the architecture of KAN(2,3,1), which includes 2 input nodes, 3 hidden nodes, and 1 output node. The output of each node is derived from the sum of individual functions \(\phi\), called "edges." The diagram also explains the computation of the inner function \(\phi\) using control points (coefficients) and B-splines. The number of B-splines is determined by adding the grid size \(G\) and the spline order \(k\), resulting in \( G + k = 3 + 3 = 6 \), corresponding to the range of \(i\) from 0 to 5.

\textbf{EfficientKAN} adopted the same approach as \citet{liu2024kan} but reworked the computation using B-splines followed by linear combination, reducing memory cost and simplifying computation\footnote{https://github.com/Blealtan/efficient-kan}. The authors replaced the incompatible L1 regularization on input samples with L1 regularization on weights. They also added learnable scales for activation functions and switched the base weight and spline scaler matrices to Kaiming uniform initialization, significantly improving performance on MNIST.

\textbf{FastKAN} can speed up training over EfficientKAN by using GRBFs to approximate the 3-order B-spline and employing layer normalization to keep inputs within the RBFs' domain~\cite{li2024kolmogorov}. These modifications simplify the implementation without sacrificing accuracy. The RBF has the formula:
\begin{equation}
\begin{aligned}
\phi(r) = e^{-\epsilon r^2}
\end{aligned}
\label{eq:gaussian_rbf}
\end{equation}
where $r = \|x - c\|$  is the distance between an input vector $x$ and a center $c$, and $\epsilon$ ($epsilon > 0$) is a sharp parameter that controls the width of the Gaussian function. FastKAN uses a special form of RBFs, Gaussian RBFs where $\epsilon = \frac{1}{2h^2}$ as~\cite{li2024kolmogorov}:

\begin{equation}
\begin{aligned}
\phi_{\mathit{RBF}}(r) = \exp\left(-\frac{r^2}{2h^2}\right)
\end{aligned}
\label{eq:special_gaussian_rbf}
\end{equation}
and $h$ for controlling the width of the Gaussian function. Finally, the RBF network with $N$ centers can be shown as~\cite{li2024kolmogorov}:

\begin{equation}
\begin{aligned}
RBF(x) = \sum_{i=1}^{N} w_i \phi_{\mathit{RBF}}(r_i) = \sum_{i=1}^{N} w_i \exp\left(-\frac{||x - c_i||}{2h^2}\right)
\end{aligned}
\label{eq:rbf_network}
\end{equation}
where $w_i$ represents adjustable weights or coefficients, and $\phi_{\mathit{RBF}}$ denotes the radial basis function as in \Cref{eq:gaussian_rbf}.

\textbf{FasterKAN} outperforms FastKAN in both forward and backward processing speeds~\cite{athanasios2024}. It uses Reflectional Switch Activation Functions (RSWAFs), which are variants of RBFs. RSWAFs are activation functions that are easy to compute because of their uniform grid structure. The RSWAF function is shown as follows:

\begin{equation}
\begin{aligned}
\phi_{\mathit{RSWAF}}(r) = 1 - \left(\tanh\left(\frac{r}{h}\right)\right)^2
\end{aligned}
\label{eq:rswaf_funct}
\end{equation}
where $h$ controls the function width. Then, the RSWAF network with $N$ centers will be:
\begin{equation}
\begin{aligned}
\mathit{RSWAF}(x) &= \sum_{i=1}^{N} w_i \phi_{\mathit{RSWAF}}(r_i) \\ &= \sum_{i=1}^{N} w_i \left(1 - \left(\tanh\left(\frac{||x - c_i||}{h}\right)\right)^2\right)
\end{aligned}
\label{eq:rswaf_network}
\end{equation}

\textbf{BSRBF-KAN} is a KAN that combines B-splines from EfficientKAN and Gaussian RBFs from FastKAN in each network layer by additions~\cite{ta2024bsrbf}. It has a speedy convergence compared to EfficientKAN, FastKAN, and FasterKAN in training data.  The BSRBF function is represented as:

\begin{equation}
\begin{aligned}
\phi_{BSRBF}(x)  =  w_b b(x) + w_s (\phi_{BS}(x) + \phi_{RBF}(x)) 
\end{aligned}
\label{eq:bsrbf_funct}
\end{equation}
where $b(x)$ and $w_b$ are the base function (linear) and its base matrix. $\phi_{BS}(x)$ and $\phi_{RBF}(x)$ are B-spline and RBF functions, and $w_s$ is the common spline matrix for both functions. 

\textbf{Wav-KAN} is a neural network architecture that integrates wavelet functions into Kolmogorov-Arnold Networks to address challenges in interpretability, training speed, robustness, and computational efficiency found in MLP and LiuKAN~\cite{bozorgasl2024wav}. By efficiently capturing both high and low-frequency components of input data, Wav-KAN achieves a balance between accurately representing the data structure and avoiding overfitting. The authors used several wavelet types, including the DoG, Mexican hat, Morlet, and Shannon. In our paper, we use the DoG function to combine other functions to create function combinations. The formula for DoG is:

\begin{equation}
\begin{aligned}
\psi(x) = \phi_{DOG}(x) = - \frac{d}{dx} \left( e^{-\frac{x^2}{2}} \right) 
= x \cdot e^{-\frac{x^2}{2}}
\end{aligned}
\label{eq:dog}
\end{equation}
which $\frac{d}{dx}$ is used to represent the derivative with respect to $x$. The term inside the derivative, \( e^{-\frac{x^2}{2}} \), is a Gaussian function centered at 0. 

For the simulation of function plots, we present them in \Cref{fig:func_plot}, where they are fitted to pass through four selected points. These plots provide a visual representation of the different function types and how they interact with the data points, helping to illustrate their shapes, which are not intended for pre-validating model performance trained on these functions. The functions analyzed include B-spline (3rd-degree), DoG, GRBF, RSWAF, and BSRBF (a combination of B-spline and GRBF). Among these, only the B-spline perfectly passes through all points, while DoG, BSRBF, and GRBF intersect with one point each, and RSWAF passes through none. It is important to note that a function passing through all specified points does not guarantee strong model performance in neural networks, as issues such as overfitting, poor generalization, and sensitivity to outliers may arise.

\subsection{FC-KAN}\

\citet{ta2024bsrbf} introduced the idea of combining functions, such as B-splines and GRBFs in BSRBF-KAN, to improve convergence when training models for image classification. However, their method was limited to element-wise addition of function outputs in each layer, without exploring other matrix operations like multiplications or different combinations. We argue that this approach might not effectively capture the input data's features. It is important to note that multiplying matrices of high-dimensional data can lead to memory errors or inefficient running time on GPU/CPU devices. Therefore, it is wise to perform these operations on low-dimensional data, such as the output layer of a neural network, in data classification problems.

\begin{figure*}[htbp]
  \centering
\includegraphics[scale=1]{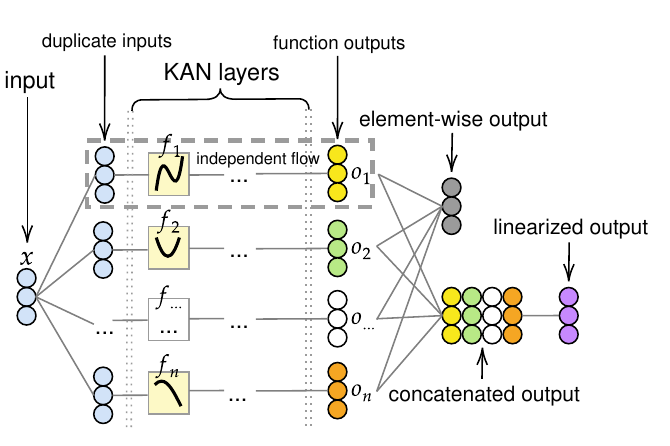}
  \centering
  \caption{The structure of FC-KAN and the three types of combined outputs: element-wise, concatenation, and linearization.}
\label{fig:fc_kan_diagram}
\end{figure*}

\begin{figure*}[htbp]
  \centering
\includegraphics[scale=1]{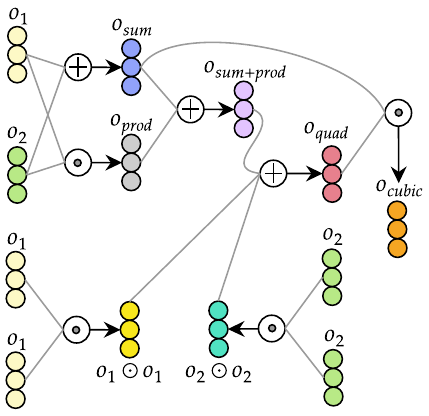}
  \centering
  \caption{Various data combinations are performed using element-wise operations (additions \(+\) and multiplications \(\odot\)) over two given outputs. The outputs always have the same data dimensions as the inputs.}
\label{fig:element_wise_outputs}
\end{figure*}

We propose a novel network, FC-KAN (Function Combinations in Kolmogorov-Arnold Networks), which leverages function combinations applied to training data, considering the outputs as low-dimensional data. Given an input \(\mathbf{x}\) and a set of functions \(F = \{f_1, f_2, \dots, f_n\}\), where \(n\) is the number of functions used, the input \(\mathbf{x}\) is passed independently to each function \(f_i\) through network layers, producing the output \(\mathbf{o_i}\) as:

\begin{equation}
\begin{aligned}
\mathbf{o_{i}} = f_i(\mathbf{x}) = (f_{i,L-1} \circ f_{i,L-2} \circ \cdots \circ f_{i,1} \circ f_{i,0})\mathbf{x}
\end{aligned}
\label{eq:single_output}
\end{equation}
which $f_{i,l}$ is the function $f_i$ at the layer $l$. So we have a set of outputs $O = \{\mathbf{o_1}, \mathbf{o_2},...,\mathbf{o_n}\}$ corresponding to the number of functions used. Note that all outputs in $O$ have the same size. 

After that, we combine function outputs to obtain the element-wise output, the concatenated output, and the linearized output, as shown in \Cref{fig:fc_kan_diagram}. The element-wise output is formed by using element-wise operations, including the sum output (\Cref{eq:output_sum}), the element-wise product output (\Cref{eq:output_product}), the sum and element-wise product output (\Cref{eq:output_sum_product}), the quadratic and cubic function outputs (\Cref{eq:output_quadratic_output} and \Cref{eq:output_cubic_output}). The concatenated output is formed by simply concatenating the function outputs together as in \Cref{eq:concat_output}. The linearized output is formed by passing the concatenated output through a linear transformation (with matrix multiplication by \( W \) and addition of bias \( b \)) as in \Cref{eq:linear_output}. The element-wise output and the linearized output will maintain the same data dimension as each element $\mathbf{o_i}$ in $O$, except in the concatenated output, where the data size will be  $\mathbf{o_i}$ multiplied by the number of function outputs being combined.

\begin{subequations}
\begin{equation}
  \mathbf{o}_{sum} = \sum_{i=1}^{n}\mathbf{o}_i  = \mathbf{o}_1 + \mathbf{o}_2 + \cdots + \mathbf{o}_n
  \label{eq:output_sum}
\end{equation}    
\begin{equation}
  \mathbf{o}_{prod} = \bigodot_{i=1}^{n}\mathbf{o}_i = \mathbf{o}_1 \odot \mathbf{o}_2 \odot \cdots \odot \mathbf{o}_n
  \label{eq:output_product}
\end{equation}
\begin{equation}
  \mathbf{o}_{sum+prod} = \mathbf{o}_{sum} + \mathbf{o}_{prod} = \sum_{i=1}^{n}\mathbf{o}_i + \bigodot_{i=1}^{n}\mathbf{o}_i 
  \label{eq:output_sum_product}
\end{equation}
\begin{equation}
\begin{split}
\mathbf{o}_{quad} 
= &\mathbf{o}_{sum+prod} + \sum_{i=1}^{n} \mathbf{o_i}\odot\mathbf{o_i} \\
= & \mathbf{o}_{sum+prod} + \mathbf{o}_1 \odot \mathbf{o}_1 + \mathbf{o}_2 \odot \mathbf{o}_2 + \dots + \mathbf{o}_n \odot \mathbf{o}_n
\end{split}
\label{eq:output_quadratic_output}
\end{equation}

\begin{equation}
\begin{split}
\mathbf{o}_{cubic} = \mathbf{o}_{quad} \odot \mathbf{o}_{sum}
\end{split}
\label{eq:output_cubic_output}
\end{equation}

\begin{equation}
  \mathbf{o}_{concat}
  = concat(\mathbf{o}_1, \mathbf{o}_2, \cdots, \mathbf{o}_n)
  \label{eq:concat_output}
\end{equation}

\begin{equation}
  \mathbf{o}_{concat\_linear} = W \cdot \mathbf{o}_{concat} + b
  \label{eq:linear_output}
\end{equation}

\end{subequations}

\Cref{fig:element_wise_outputs} presents data combinations over two given function outputs $\mathbf{o_1}$ and $\mathbf{o_2}$ using element-wise operations. The results are 5 combined outputs: sum, sum + prod, prod, quad, and cubic. Output combinations can utilize higher-degree functions, but these may significantly increase computational complexity, especially in matrix multiplication. Additionally, using more functions results in a larger number of outputs, which can further complicate data combination calculations. To manage this complexity, we prefer to restrict output combinations to quadratic functions involving up to two outputs. For instance, to combine DoG and B-splines at the output, we can use the following quadratic function formula:

\begin{equation}
\begin{aligned}
  \mathbf{o}_{DoG+BS} &= f_{DoG}(\mathbf{x}) + f_{BS}(\mathbf{x}) + f_{DoG}(\mathbf{x}) \odot f_{BS}(\mathbf{x}) +  (f_{DoG}(\mathbf{x}))^2 + (f_{BS}(\mathbf{x}))^2 \\
  &= \cdots + \cdots + \cdots + f_{DoG}(\mathbf{x}) \odot f_{DoG}(\mathbf{x}) + f_{BS}(\mathbf{x}) \odot f_{BS}(\mathbf{x}) \\
  &= \mathbf{o}_{DoG} + \mathbf{o}_{BS} + \mathbf{o}_{DoG} \odot \mathbf{o}_{BS} + \mathbf{o}_{DoG} \odot \mathbf{o}_{DoG} + \mathbf{o}_{BS} \odot \mathbf{o}_{BS}
\end{aligned}
\label{eq:dog_bs_quadratic}
\end{equation}
which $f_{DoG}$ and $f_{BS}$ refer to DoG and B-spline functions. Finally, we use the combined output to compute the cross-entropy loss against the true labels when training the models.

\section{Experiments}
\subsection{Training Configuration}
There are 5 independent training runs for each model on the MNIST~\cite{deng2012mnist} and Fashion-MNIST~\cite{xiao2017fashion} datasets to obtain a more reliable overall performance assessment. We then calculate the average value from all runs to minimize the impact of training variability and accurately gauge the models' maximum potential. To maintain simplicity in the network design, we utilized only activation functions (SiLU), linear transformations, and layer normalization in all models: BSRBF-KAN, EfficientKAN, FastKAN, FasterKAN, FC-KAN, and MLP. We do not use LiuKAN because its design, as the author intended, results in longer training times~\cite{liu2024kan}.

\begin{table*}[ht]
	\caption{The number of parameters by models. The number of parameters includes both used and unused parameters.}
	\centering
	\begin{tabular}{p{4cm}p{2.5cm}p{3cm}p{2cm}}
            \hline
		\textbf{Dataset} & \textbf{Model} & \textbf{Network structure} & \textbf{\#Params} \\
    \hline 
    \multirow{6}{4cm}{\textbf{MNIST + Fashion-MNIST}}  &  BSRBF-KAN	& (784, 64, 10) & 459040  \\
&  FastKAN	& (784, 64, 10) & 459114 \\
& FasterKAN	& (784, 64, 10) & 408224 \\
& EfficientKAN	& (784, 64, 10) & 508160 \\
& FC-KAN &	(784, 64, 10) & \textbf{560820} \\
& MLP & (784, 64, 10) & 52512 \\
            \hline
            
             \hline
	\end{tabular}
	\label{tab:model_params}
\end{table*}

\begin{figure*}[htbp]
  \centering

\includegraphics[scale=0.75]{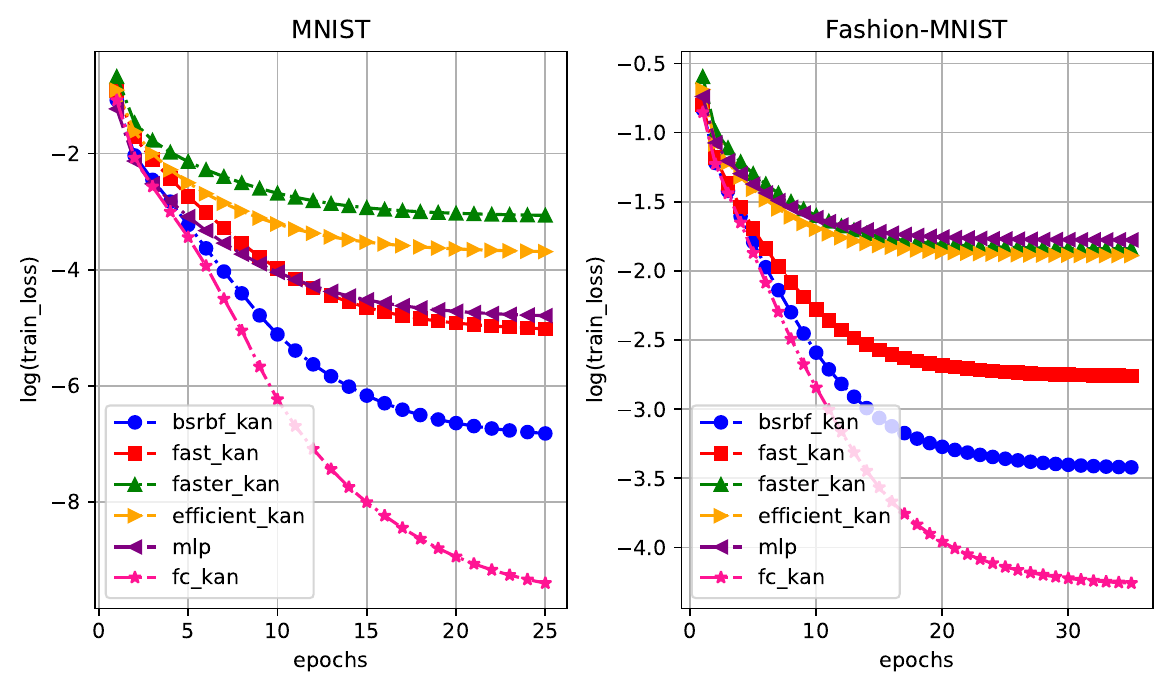}

  \centering
  \caption{The logarithmic values of training losses for the models over 25 epochs on MNIST and 35 epochs on Fashion-MNIST. A quadratic function is used to combine B-splines and DoG at the output of FC-KAN.}
\label{fig:train_losses}
\end{figure*}

As shown in \Cref{tab:model_params}, all models contain a network structure of (784, 64, 10), comprising 784 input neurons, 64 hidden neurons, and 10 output neurons corresponding to the 10 output classes (0-9). Due to the function combinations, FC-KAN has the highest number of parameters, while the MLP has the fewest because it only contains linear transformations over data between layers. The models were trained with 25 epochs on MNIST and 35 epochs on Fashion-MNIST. For KAN models, we use  \texttt{grid\_size=5}, \texttt{spline\_order=3}, and \texttt{num\_grids=8}. Other hyperparameters are the same in all models, including \texttt{batch\_size=64}, \texttt{learning\_rate=1e-3}, \texttt{weight\_decay=1e-4}, \texttt{gamma=0.8}, \texttt{optimize=AdamW}, and \texttt{loss=CrossEntropy}.

In FC-KAN models, we combine 2 out of 4 functions: B-splines (denoted as BS), Radial Basis Functions (denoted as RBF, specifically using GRBFs), \textbf{color}{Derivative of Gaussians} (denoted as DoG), and linear transformations (denoted as BASE), to create 6 FC-KAN variants. All variants use a quadratic function representation in the output for the experiments. The FC-KAN models are: FC-KAN (DoG+BS), FC-KAN (DoG+RBF), FC-KAN (DoG+BASE), FC-KAN (BS+RBF), FC-KAN (BS+BASE), and FC-KAN (RBF+BASE).

\subsection{Model Performance}
\begin{table*}[ht]
	\caption{The average metric values in 5 training runs on MNIST and Fashion-MNIST. FC-KAN models use a quadratic function representation to combine outputs.}
	\centering
	\begin{tabular}{p{2cm}p{3.7cm}p{2cm}p{2cm}p{2cm}p{1.4cm}}
            \hline
		\textbf{Dataset} &  \textbf{Model}   & \textbf{Train. Acc.} & \textbf{Val. Acc.} & \textbf{F1} & \textbf{Time (seconds)} \\
    \hline 
	\multirow{6}{2cm}{\textbf{MNIST}} &	BSRBF-KAN	& \textbf{100.00 ± 0.00} & 97.59 ± 0.02 & 97.56 ± 0.02 & 211.5 \\
& FastKAN	& 99.98 ± 0.01 & 97.47 ± 0.05 & 97.43 ± 0.05 & 164.47 \\
& FasterKAN	 & 98.72 ± 0.02 & 97.69 ± 0.04 & 97.66 ± 0.04 & 161.88 \\
& EfficientKAN	&	99.40 ± 0.10 & 97.34 ± 0.05 & 97.30 ± 0.05 & 184.5 \\
& MLP & 99.82 ± 0.08 & 97.74 ± 0.07 & 97.71 ± 0.07 & \textbf{146.58} \\
& \textbf{FC-KAN (DoG+BS)} & \textbf{100.00 ± 0.00} & 97.91 ± 0.05 & 97.88 ± 0.05 & 263.29 \\
& FC-KAN (DoG+RBF) & \textbf{100.00 ± 0.00} & 97.76 ± 0.04 &  97.73 ± 0.04 & 225.23 \\
& FC-KAN (DoG+BASE) & 99.82 ± 0.11 &  97.76 ± 0.02 &  97.73 ± 0.02 & 213.87 \\
&  FC-KAN (BS+RBF) &  99.99 ± 0.00  & 97.53 ± 0.04 & 97.49 ± 0.04 & 233.89 \\
& FC-KAN (BS+BASE) &  \textbf{100.00 ± 0.00} & \textbf{97.93 ± 0.05} & \textbf{97.91 ± 0.05} & 238.94 \\
& FC-KAN (RBF+BASE) & \textbf{100.00 ± 0.00} & 97.85 ± 0.03 &  97.82 ± 0.04 &  193.71 \\
            \hline
            \hline
           \multirow{6}{2cm}{\textbf{Fashion\\-MNIST}} & BSRBF-KAN	&  99.34 ± 0.04 & 89.38 ± 0.06 & 89.36 ± 0.06 & 276.75  \\
& FastKAN	&	98.25 ± 0.07 & 89.40 ± 0.08 & 89.35 ± 0.08 & 208.68 \\
& FasterKAN	&	94.41 ± 0.03 & 89.31 ± 0.03 & 89.25 ± 0.02 & 220.7 \\
& EfficientKAN	&	94.81 ± 0.09 & 88.98 ± 0.07 & 88.91 ± 0.08 & 247.85 \\
& MLP & 94.14 ± 0.04 & 88.94 ± 0.05 & 88.88 ± 0.05 & \textbf{200.28} \\
& \textbf{FC-KAN (DoG+BS)} & 99.54 ± 0.13 & \textbf{89.99 ± 0.09} & \textbf{89.93 ± 0.08} & 369.2 \\
& FC-KAN (DoG+RBF) & \textbf{99.82 ± 0.03} & 89.86 ± 0.12 & 89.81 ± 0.12 & 309.81 \\
& FC-KAN (DoG+BASE) & 95.36 ± 0.13  &  89.57 ± 0.07 & 89.49 ± 0.07 & 300.13 \\
& FC-KAN (BS+RBF) & 99.60 ± 0.09 & 89.45 ± 0.10  & 89.43 ± 0.10 & 330.82 \\
& FC-KAN (BS+BASE) & 99.73 ± 0.02  & 89.90 ± 0.09 &  89.85 ± 0.10 & 326.57 \\
& FC-KAN (RBF+BASE) & 99.79 ± 0.03 &  89.69 ± 0.03 &  89.65 ± 0.04 & 277.13 \\
            \hline
             \multicolumn{6}{l}{Train. Acc = Training Accuracy, Val. Acc. = Validation Accuracy }  \\
             \multicolumn{6}{l}{BASE = linear transformations, BS = B-splines, DoG = \textcolor{blue}{Derivative of Gaussians}, RBF = Radial Basis Functions}  \\
             \hline
	\end{tabular}
	\label{tab:average_metric}
\end{table*}

\Cref{fig:train_losses} shows the training losses, represented on a logarithmic scale, for MLP and KAN models on the MNIST and Fashion-MNIST datasets. The loss performance of each model was evaluated based on an independent training run. FC-KAN consistently achieves the lowest training losses across both datasets, followed by BSRBF-KAN due to its fast convergence feature. In contrast, FasterKAN records the highest training loss on MNIST, while MLP performs similarly on Fashion-MNIST.

In general, FC-KAN models outperformed others on MNIST and Fashion-MNIST but require more training time due to the quadratic function representation for output combination, as shown in \Cref{tab:average_metric}. This trade-off between training time and model performance is considered reasonable. The best-performing models are FC-KAN (BS+BASE) and FC-KAN (DoG+BS), which achieved validation accuracies of 97.93\% on MNIST and 89.99\% on Fashion-MNIST, respectively. When calculating the metric values for both datasets, FC-KAN (DoG+BS) slightly surpassed FC-KAN (BS+BASE) and outperformed the other models. However, FC-KAN (BS+BASE) models take between 9.25\% and 11.55\% less training time. 

Although it has the lowest performance on Fashion-MNIST, the MLP model has the fastest training time and demonstrates competitive accuracy on MNIST, even outperforming BSRBF-KAN, FastKAN, FasterKAN, and EfficientKAN on this dataset. On MNIST, MLP also contributes to the success of FC-KAN (BS+BASE), which combines outputs of B-splines and linear transformations.

BSRBF-KAN, FC-KAN (DoG+BS), FC-KAN (DoG+RBF), FC-KAN (BS+BASE), and FC-KAN (RBF+BASE) exhibit the best convergence on MNIST, while FC-KAN (DoG+RBF) performs the best on Fashion-MNIST, followed by FC-KAN (RBF+BASE) and FC-KAN (BS+BASE). We observe that fast convergence is achieved in KAN models that incorporate function combinations rather than relying on single functions. This finding is important to consider when designing KANs with a focus on achieving rapid convergence.

\subsection{Comparison of Combination Methods}
\begin{table*}[ht]
	\caption{The performance of FC-KAN (DoG+BS) using different output combination methods.}
	\centering
	\begin{tabular}{p{2cm}p{3.2cm}p{2cm}p{2cm}p{2cm}p{2.5cm}}
            \hline
		\textbf{Dataset} & \textbf{Combined Method} & \textbf{Train. Acc.} & \textbf{Val. Acc.} & \textbf{F1} & \textbf{Time (seconds)} \\
    \hline 
	\multirow{6}{2cm}{\textbf{MNIST}} & Sum & \textbf{100.00 ± 0.00} & 97.61 ± 0.04 & 97.58 ± 0.04 & 247.12 \\
& Product & \textbf{100.00 ± 0.00} & 97.59 ± 0.07 & 97.56 ± 0.07 & 247.5 \\
& Sum + Product & \textbf{100.00 ± 0.00} & 97.73 ± 0.04 & 97.70 ± 0.04 & \textbf{244.95} \\
& Quadratic Function & \textbf{100.00 ± 0.00} & \textbf{97.91 ± 0.05} & \textbf{97.88 ± 0.05} & 263.29 \\
& Cubic Function & \textbf{100.00 ± 0.00} & 97.68 ± 0.05 & 97.65 ± 0.05 & 266.46 \\
& Concatenation &  99.64 ± 0.07 & 97.20 ± 0.02 & 97.16 ± 0.02 & 250.03 \\
& Linear Concatenation & \textbf{100.00 ± 0.00} & 97.60 ± 0.04 & 97.56 ± 0.05 & 253.88 \\
& Minimum & 99.92 ± 0.04 & 97.23 ± 0.04 & 97.20 ± 0.04 & 253.21 \\
& Maximum & 99.97 ± 0.01 & 97.30 ± 0.05 & 97.26 ± 0.05 & 249.56 \\
& Average & \textbf{100.00 ± 0.00} & 97.44 ± 0.02 & 97.40 ± 0.03 & 255.4 \\
            \hline
            \hline
           \multirow{6}{2cm}{\textbf{Fashion\\-MNIST}} & Sum & 99.39 ± 0.03 & 89.56 ± 0.07 & 89.55 ± 0.09 & 346.21 \\
& Product & 99.50 ± 0.05 & 89.95 ± 0.08 & 89.90 ± 0.08 & 345.85 \\
& Sum + Product & \textbf{99.56 ± 0.05} & 89.89 ± 0.13 & 89.84 ± 0.13 & 349.4 \\
& Quadratic Function & 99.54 ± 0.13 & \textbf{89.99 ± 0.09} & \textbf{89.93 ± 0.08} & 369.2 \\
& Cubic Function & 99.40 ± 0.05 & 89.69 ± 0.10 & 89.67 ± 0.09 & 367.83 \\
& Concatenation & 95.27 ± 0.05 & 89.09 ± 0.04 & 89.01 ± 0.04 & \textbf{345.35} \\
& Linear Concatenation & 99.53 ± 0.03 & 89.40 ± 0.06 & 89.37 ± 0.07 & 358.16 \\
& Minimum & 98.13 ± 0.27 & 89.37 ± 0.06 & 89.33 ± 0.06 & 351.53 \\
& Maximum & 97.47 ± 0.68 & 89.34 ± 0.06 & 89.28 ± 0.06 & 353.33 \\
& Average & 99.09 ± 0.03 & 89.54 ± 0.06 & 89.51 ± 0.06 & 354.74 \\
            \hline
             \multicolumn{6}{l}{Train. Acc = Training Accuracy, Val. Acc. = Validation Accuracy }  \\
             \hline
	\end{tabular}
	\label{tab:combination_types}
\end{table*}

While employing a quadratic function representation for the output of FC-KAN, we are also interested in exploring how different output combination methods affect model performance. In this experiment, we use FC-KAN (DoG+BS) with several output combination methods: sum, element-wise product, addition of sum and element-wise product, quadratic and cubic function representations,  concatenation, and linear transformation of the concatenated output. Inspired by the work of \citet{altarabichi2024rethinking} on DropKAN~\cite{altarabichi2024dropkan}, we also include the maximum, minimum, and average outputs for comparison.

From the results in \Cref{tab:combination_types}, the quadratic function best represents the combination output and outperforms other combinations, although its models require more training time. It is clear that the outputs combined by element-wise operations consistently outperform other methods, demonstrating superior accuracy. This indicates that element-wise combinations are more effective in capturing and integrating relevant features from the data, leading to better performance. Meanwhile, the output combination by concatenation shows the worst results. The linear transformation applied to the concatenated output outperforms concatenation alone, but still achieves only average performance. Similarly, the maximum, minimum, and average outputs do not deliver superior results.

In MNIST, besides the quadratic functions, the addition of sum and element-wise product demonstrates very competitive performance while requiring the least training time. Except for concatenation, maximum, and minimum, all other combinations can easily achieve 100\% training accuracy. In Fashion-MNIST, the element-wise product combination is only surpassed by the quadratic function, but the plus point is it takes 6.3\% less training time. The addition of sum and element-wise product achieves the best training accuracy, followed by the quadratic function and the linear transformation of concatenation.

Contrary to our expectations, the cubic function representation only achieves moderate performance. However, it takes the longest training time on MNIST and ranks just behind the quadratic function in terms of training time on Fashion-MNIST. Initially, we hypothesized that the cubic representation could capture more data features, but it appears that the excessive number of element-wise operations may hinder feature extraction, potentially leading to reduced performance. This experiment demonstrates that using higher-degree functions may not necessarily enhance model performance and can also increase computational complexity.

\subsection{Misclassification Analysis}
\begin{figure*}[htbp]
  \centering

\includegraphics[scale=0.55]{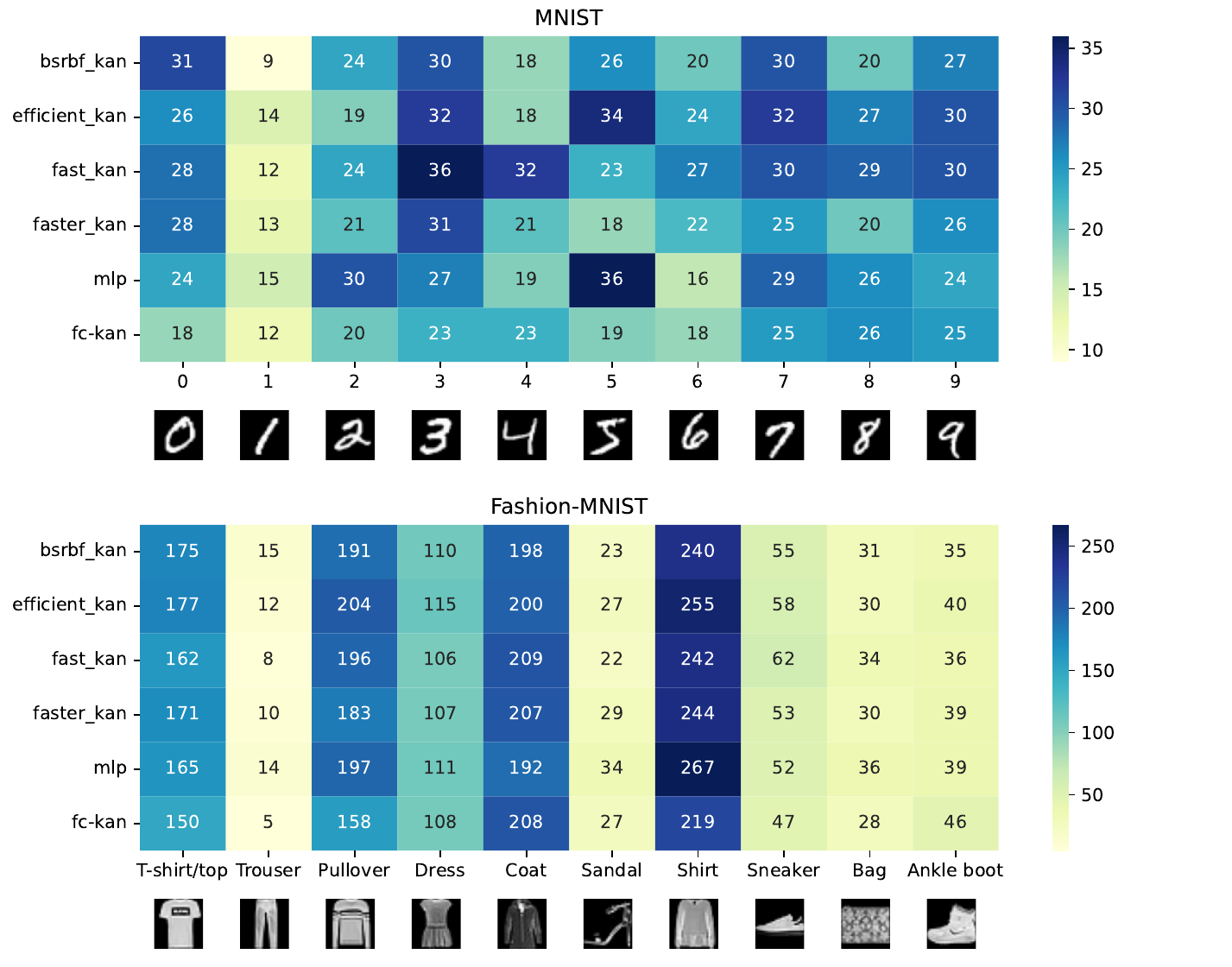}

  \centering
  \caption{Heatmaps of the misclassified images in the validation set by models over MNIST and Fashion-MNIST. FC-KAN (DoG+BS) is used for comparison and has the fewest total errors compared to other models.}
\label{fig:misclass}
\end{figure*}

To evaluate model performance across classes, we conducted a qualitative analysis of misclassifications on the validation set of the MNIST and Fashion-MNIST datasets. We selected FC-KAN (DoG+BS) along with other KANs for comparison, with each model trained for only 1 run. For each model, we counted the raw frequency of misclassified images per output class. Both MNIST and Fashion-MNIST have 10 output classes. In MNIST, the classes range from 0 to 9, while in Fashion-MNIST, they include "T-shirt/top", "Trouser", "Pullover", "Dress", "Coat", "Sandal", "Shirt", "Sneaker", "Bag", and "Ankle boot".

In MNIST, FC-KAN generally exhibits the fewest misclassification errors by class, while other models show their own weaknesses. For example, MLP and EfficientKAN performed poorly on Class 5, FastKAN on Classes 3 and 4, and BSRBF-KAN on Class 0. Class 1 had the fewest errors, while all models struggled with certain classes, such as Classes 3, 7, and 9. It is surprising in the case of Class 7, as its images seemed easy to recognize from our perspective.

In Fashion-MNIST, the performance of models by class is generally similar, but FC-KAN still outperforms in some classes, such as "T-shirt/top", "Pullover", and "Shirt." Models perform very well in certain classes, such as "Trouser", "Sandal", "Sneaker", "Bag", and "Ankle boot", while they struggle more with recognizing images belonging to "T-shirt/top", "Pullover", "Coat", and "Shirt" due to their similar appearance. This can be referred to as classification ambiguity due to the nature of the data. 

In short, FC-KAN outperforms other KANs, but it only mitigates, not completely resolves, the misclassification errors found in the other models. This analysis is also helpful for focusing on design methods to address the most challenging classes and improve recognition accuracy.

\subsection{Model Performance with Limited Data}

\begin{figure*}[htbp]
  \centering
\includegraphics[scale=0.5]{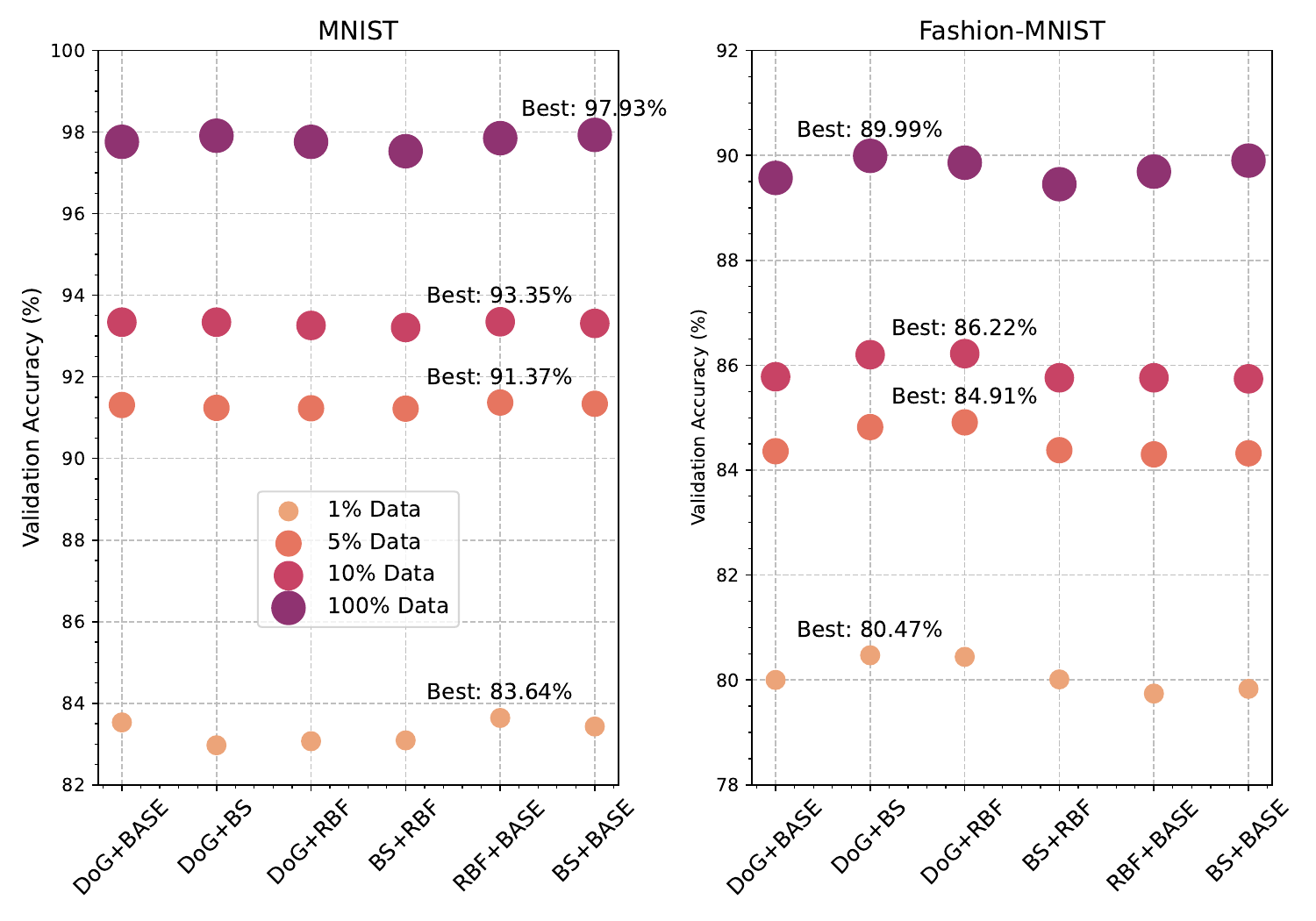}
  \centering
  \caption{The validation accuracy values of the models across various data subsets. While achieving the highest accuracy values with full data training on MNIST, FC-KAN (BS+BASE) shows no significant advantage with lesser training data. Similarly, FC-KAN (DoG+BS) performs optimally only with 1\% of the data.}
\label{fig:subset_plot}
\end{figure*}

Rather than training FC-KAN models on the full data to determine the optimal configuration, we investigate whether using smaller portions of the data can yield comparable insights. This approach not only helps identify the best function combinations but also significantly reduces training time. Additionally, it allows for testing a greater number of configurations and provides early indications of model performance on the full training data. In this experiment, we evaluate 6 FC-KAN variants—FC-KAN (DoG+BS), FC-KAN (DoG+RBF), FC-KAN (DoG+BASE), FC-KAN (BS+RBF), FC-KAN (BS+BASE), and FC-KAN (RBF+BASE)—using 1\%, 5\%, and 10\% of the data, with a quadratic function representation for combining function outputs. Each model is trained over 5 independent runs with the same configuration, and we calculate the average performance.

\Cref{fig:subset_plot} illustrates the performance of the FC-KAN models across different training subsets and function combinations. In the MNIST dataset, the RBF+BASE combination demonstrates superior performance with 1\%, 5\%, and 10\% of the data; however, the highest performance on the full training data is recorded for the BS+BASE combination. This observation underscores the challenges of predicting the optimal function combination when working with limited training data. In the Fashion-MNIST dataset, DoG+BS achieves the best performance with 1\% of the data, while DoG+RBF excels with both 5\% and 10\%. These results suggest that DoG+BS and DoG+RBF may perform well with the full training dataset. Indeed, DoG+BS consistently outperforms other combinations in full training, with DoG+RBF serving as a strong competitor, as in \Cref{tab:average_metric}. Overall, our findings indicate that accurately predicting the performance of FC variants on the complete training dataset based solely on their performance with smaller subsets is challenging. This variability may depend on the specific datasets and the portions of data used for training.

\section{Limitation}
Although FC-KAN is designed to utilize data combinations in low-dimensional layers, our experiments applied it only to the output layer, considered a low-dimensional layer in a network with the structure (784, 64, 10). As a result, the impact of these combinations on model performance in deeper network architectures with low-dimensional layers remains unclear. Another limitation is the number of parameters in the models. In the experiments, the MLP used the fewest parameters within the same network structure (784, 64, 10) compared to other models. We are also interested in how MLP would perform relative to KAN models if they have the same number of parameters. According to \citet{yu2024kan}, MLP generally outperforms KAN models, except in tasks involving symbolic formula representation.

We also question whether the model's performance would improve if data combinations were applied in all layers, rather than just low-dimensional layers, assuming that device memory constraints are not an issue in data multiplications. Finally, since FC-KAN has only been tested on two datasets, MNIST and Fashion-MNIST, more datasets should be used to properly evaluate its effectiveness. In short, these limitations can be addressed by designing network structures that integrate low-dimensional data and evaluating them across various problems or through additional experiments for greater clarity.

\section{Conclusion}
We introduced FC-KAN, which uses various popular mathematical functions to represent data features and combines their outputs using different methods, primarily through element-wise operations in low-dimensional layers, to address image classification problems. In the experiments, we designed FC-KAN to combine pairs of functions, such as B-splines, wavelets, and radial basis functions, using several output combinations on the MNIST and Fashion-MNIST datasets. 

We found that FC-KAN outperformed MLP and other KAN models in terms of accuracy using the same network structure, although it required more training time. This is supported by a misclassification analysis, where FC-KAN achieves the fewest errors per class but still exhibits classification ambiguity errors, similar to other models. Among the variants, FC-KAN (DoG+BS) and FC-KAN (BS+BASE), which combine DoGs and B-splines, as well as MLPs and B-splines, respectively, in a quadratic function representation of the output, achieved the best results on both datasets. Our experiments also show that it is not easy to detect the function combination performance in the full training data based on small-scale experiments. 

FC-KAN has promising potential for utilizing function combinations to design KANs and enhance model performance. However, we will aim to investigate several aspects further. These include exploring alternative functions and combinations for feature extraction, developing methods to reduce parameter usage while maintaining or improving model performance, and examining the impact of different function combinations on the stability and efficiency of KANs. These will be the focus of our future work.

\bibliographystyle{unsrtnat}
\bibliography{references}  






\end{document}